\documentclass[conference, 9pt]{IEEEtran}
\IEEEoverridecommandlockouts

\usepackage{cite}
\usepackage{amsmath,amssymb,amsfonts}
\usepackage{algorithmic}
\usepackage{graphicx}
\usepackage{textcomp}
\usepackage{xcolor}
\usepackage{hyperref}
\usepackage{subfig} 
\usepackage{float}
\usepackage{booktabs}
\usepackage{soul}

\def\BibTeX{{\rm B\kern-.05em{\sc i\kern-.025em b}\kern-.08em
    T\kern-.1667em\lower.7ex\hbox{E}\kern-.125emX}}
\begin{document}

\title{Iceberg: Enhancing HLS Modeling with Synthetic Data}

\renewcommand{\baselinestretch}{0.950}

\author{\IEEEauthorblockN{Zijian Ding, Tung Nguyen, Weikai Li, Aditya Grover, Yizhou Sun, Jason Cong}
\IEEEauthorblockA{\textit{Computer Science Department, UCLA} \\
\{bradyd, tungnd, weikaili, adityag, yzsun, cong\}@cs.ucla.edu}
}

\maketitle

\begin{abstract}
Deep learning-based prediction models for High-Level Synthesis (HLS) of hardware designs often struggle to generalize. In this paper, we study how to close the generalizability gap of these models through pretraining on synthetic data and introduce Iceberg, a synthetic data augmentation approach that expands both large language model (LLM)-generated programs and weak labels of unseen design configurations. Our weak label generation method is integrated with an in-context model architecture, enabling meta-learning from actual and proximate labels. Iceberg improves the geometric mean modeling accuracy by $86.4\%$ when adapt to six real-world applications with few-shot examples and achieves a $2.47\times$ and a $1.12\times$ better offline DSE performance when adapting to two different test datasets. Our open-sourced code is here: \href{https://github.com/UCLA-VAST/iceberg}{https://github.com/UCLA-VAST/iceberg}.
\end{abstract}

\section{Introduction}
\label{sec:intro}

High-Level Synthesis (HLS)~\cite{cong2011high,cong2022fpga} was introduced to simplify the design of domain-specific accelerators (DSAs) by raising the abstraction level from the register-transfer level (RTL) design to C/C++ based HLS designs. Despite recent interests in generating RTL code with large language models (LLMs)~\cite{verilogeval, zhang2024mg, cui2024origen}, HLS remains a competitive solution for exploring correct-by-construction transformations and for prototyping efficient hardware architectures. Several HLS-based tools have been developed to further streamline this process~\cite{scalehls,streamhls,lightningsimv2,sisyphus,allo}, offering predefined hardware architecture templates that allow designers to specify key parameters. While these tools significantly reduce the complexity of hardware accelerator development, producing high-performance designs remains challenging for developers. This difficulty arises because efficiently generating hardware from C/C++ heavily depends on selecting the optimal combination of parameters, such as HLS pragmas (Appx. \ref{appx:pragma_insertion}).

\begin{figure}[h]
    \centering
    \includegraphics[width=0.85\columnwidth]{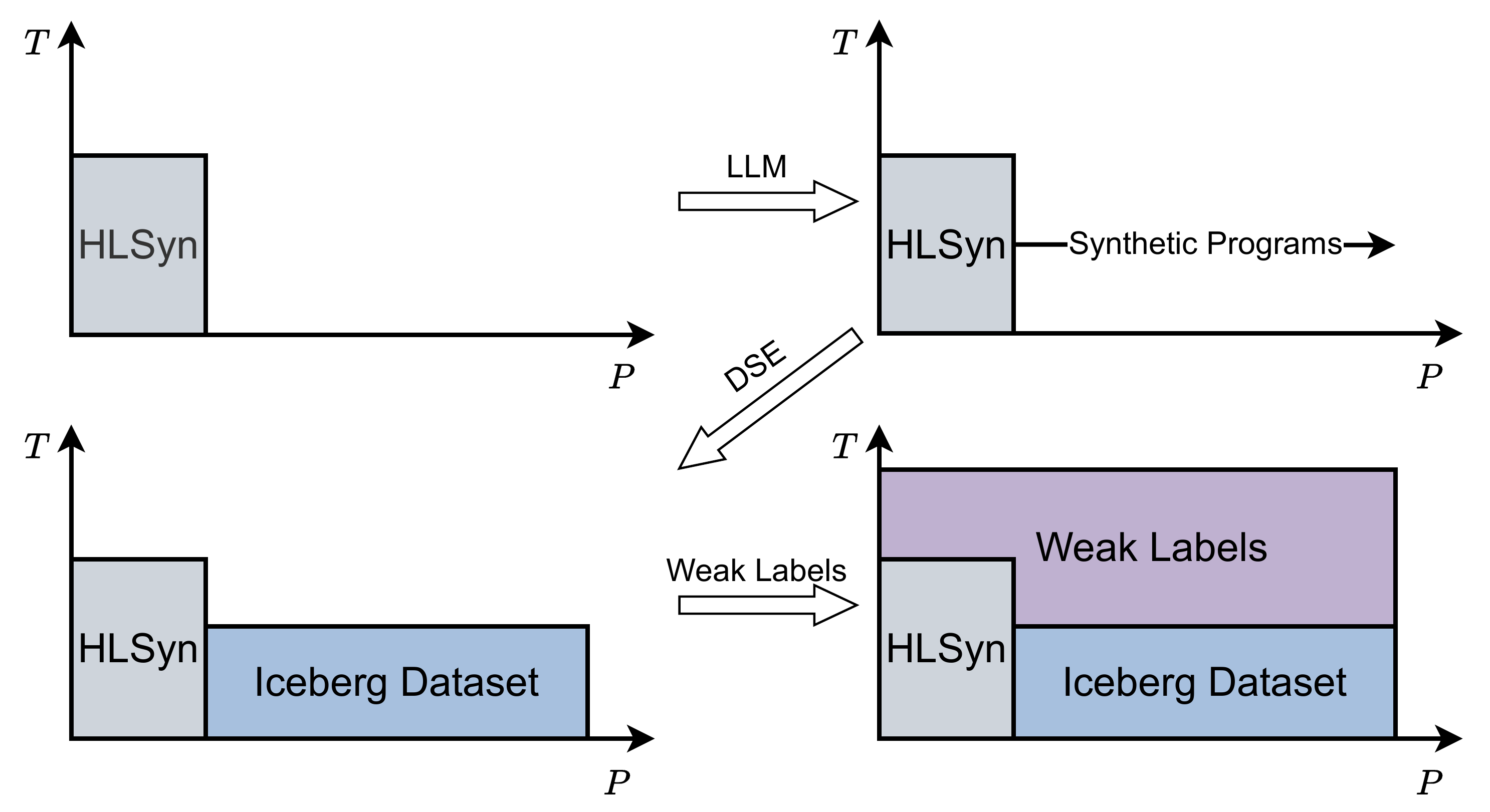}
    \caption{\small Our approach to synthetic data augmentation spans both the program (P) dimension and the design configuration (T) dimension. To expand in the P dimension, we use LLMs to generate diverse synthetic programs, then obtain an actual labeled dataset by running DSE on these programs. To scale in the T dimension, we generate weak labels for unseen T configurations. Finally, we combine actual and weak labels to form a hybrid dataset.}
    \label{fig:data}
\end{figure}

Several recent studies have explored the utilization of AI/ML to enhance the programmability of HLS and other domain-specific languages (DSLs) for accelerators~\cite{hlsdebug, hlspilot, collini2024c2hlsc,ralad,hlsrepair, llmaccel,gpt4aigchip}. Among these, model-based optimization approaches~\cite{ironmanpro,harp,murphy2024balor,sourcetopostroute,bai2024learning,powergear,qin2024cross,li2025hierarchical,active-cem} represent the state-of-the-art for pragma insertion tasks. At the core of these methods is a surrogate model designed to replace the time-consuming evaluation of downstream tools. However, these models often exhibit poor generalization capability when applied to new kernels or designs, restricting their applicability to the specific instances they were trained on~\cite{li2025hierarchical, active-cem}. The primary reason for this limitation is the scarcity of publicly available datasets for training these models. HLS designs span two key dimensions: programs (Ps) and design configurations (Ts). Unfortunately, existing data efforts~\cite{hlsyn,hlsfactory} still lack diversity in at least one of these dimensions, making it challenging to train surrogate models that can effectively generalize to unseen Ps or Ts with limited examples.

Scaling in both dimensions is challenging. In the program (P) dimension, while there are abundant publicly available C/C++ programs, many are not suitable for HLS acceleration. An ideal program should meet two key criteria: (1) Compatibility with HLS-based tools. Many tools impose strict format constraints. For instance, some tools~\cite{merlin} do not support several standard libraries. Other tools are more friendly to Python code~\cite{streamhls,allo}. (2) Suitability for customized acceleration. The program should be a suitable candidate for acceleration, which excludes inherently dynamic and sequential workloads commonly found in public datasets. A growing body of research focuses on code refactoring to convert arbitrary C/C++ code into synthesizable HLS code~\cite{collini2024c2hlsc, hlsrepair}. However, these approaches may fail to address the second requirement, as they do not necessarily ensure that the transformed code is well optimized for acceleration, frequently requiring more extensive modifications.

In the design configuration (T) dimension, the primary challenge is the high computational cost of evaluating individual designs. Synthesizing each T can take anywhere from a minute to an hour, making large-scale exploration infeasible. Additionally, due to the shortage of experienced hardware programmers, manually constructing optimal designs is impractical. One potential solution is to leverage automated tools~\cite{ye2024hida,streamhls,autodse,sisyphus}. However, identifying high-quality configurations for each P is computationally expensive, often taking hours to days.

To mitigate these challenges, we propose Iceberg, a synthetic data augmentation technique that achieves state-of-the-art few-shot learning capabilities for HLS designs. As shown in Fig. \ref{fig:data}, our method improves scalability across both program and design configuration dimensions while enhancing model performance. To scale the P dimension, we propose an LLM-driven synthetic program generator capable of producing diverse, high-quality programs for HLS (Sec. \ref{sec:iceberg}). For the T dimension, we develop a novel technique that leverages trained surrogate models to create weak labels for unseen design configurations. Then, we employ a Transformer Neural Process (TNP)~\cite{tnp} to fit both actual and weak labels (Sec. \ref{sec:tnp}). Our results reveal that the highest accuracy is attained when the model is trained with diverse weak labels generated by different synthetic functions (Appx. \ref{appx:scale_f}).

Previous works~\cite{expt,tabpfn} also examine weak labels for model-based optimization. A common approach is to utilize a large set of randomly initialized functions. However, this strategy performs poorly on HLS designs due to the mismatch between the real performance label and the randomly initialized synthetic functions. Our approach spends more compute on each synthetic function and requires a smaller set overall.

We evaluate Iceberg in three different settings. First, we show that regardless of the model architecture, pretraining with the Iceberg dataset improves generalization to the target dataset. Second, we assess the efficiency of the prediction model by comparing it against the current state-of-the-art on the HLSyn dataset~\cite{hlsyn}. Finally, we pretrain models with the Iceberg dataset and test their accuracy on the publicly available HLSyn dataset and real-world applications derived from the Rosetta benchmark~\cite{rosetta} and popular machine learning models~\cite{attention}. Across all settings, Iceberg’s performance model consistently beats existing models, achieving an 86.4\% improvement over the state-of-the-art model on real-world applications. Throughout the remaining sections, Iceberg refers interchangeably to both the synthetic data generation approach and the dataset it produces.
\section{Synthesizing diverse \textit{P} with large language models}
\label{sec:iceberg}

As stated in Sec. \ref{sec:intro}, an ideal set of Ps should consist of diverse programs that are related, ensuring a wide coverage of potential workloads. A natural approach to achieving this is through prompting LLMs. With their ability to follow instructions and acquire domain knowledge through in-context prompting, these agents can synthesize high-quality programs spanning various application domains and diverse scales. This capability enables the creation of a dataset of programs that can effectively generalize to downstream tasks.

\begin{figure}[ht]
    \centering
    \includegraphics[width=0.85\columnwidth]{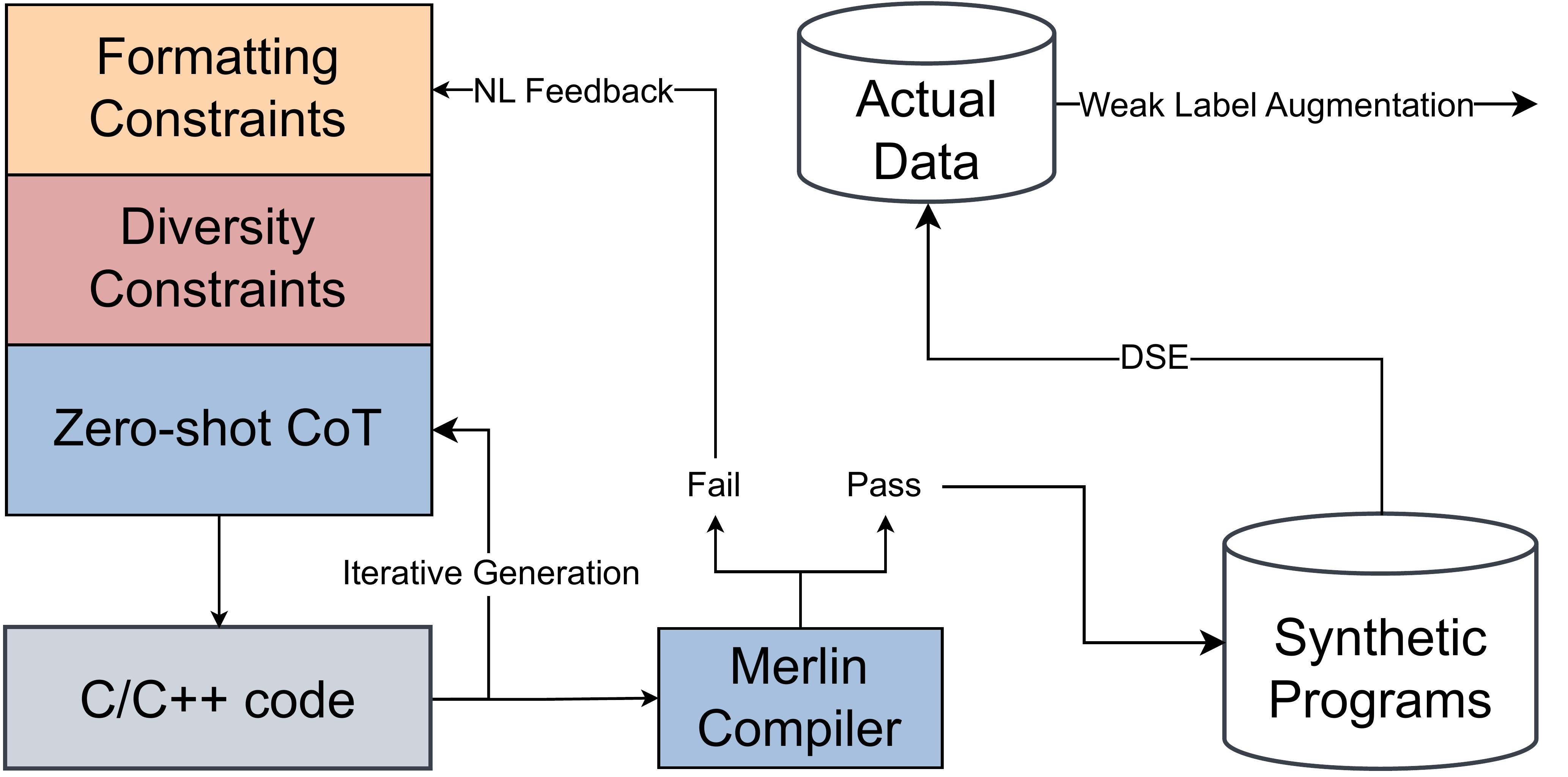}
    \caption{\small Synthesizing diverse programs with LLMs. We start with manually constructing a prompt for the target downstream tools. Beyond enforcing formatting and diversity constraints, we further apply iterative generation to construct the synthetic programs. Once the program set is established, we run design space exploration (DSE) on the generated programs to obtain design configurations with actual label. This labeled dataset then synthesizes weak labels and model training.}
    \label{fig:iceberg}
\end{figure}

Fig. \ref{fig:iceberg} presents our LLM-based framework for synthetic program generation. At its core is a dedicated initial prompt that defines all constraints and preferences for the desired program. To ensure diversity, multiple programs are iteratively generated for each application domain. After generating a batch of programs, external tools are invoked to verify their validity. Common mistakes are then summarized, and the initial prompt is refined through natural language feedback, progressively improving the quality and relevance of the generated programs.

\subsection{Constructing the initial prompt}

By manually constructing tool-specific prompts, Iceberg provides a fundamentally more scalable solution for expanding the P dimension compared to utilizing publicly available programs. Public datasets often require extensive preprocessing to meet HLS tool format constraints~\cite{vitis23}, such as removing the "main" function while preserving functionality and identifying a single entry point. However, reliably detecting the entry point is challenging, especially when programs originate from multi-file projects with complex dependencies. We prompt the LLM to directly generate C programs that omit the ``main" function. Additionally, we explicitly instruct the model to define a single entry point named ``top" and structure all input and output variables as function arguments. This approach significantly simplifies the integration of these programs with tools such as Vitis HLS~\cite{vitis23}. 

Beyond formatting, Iceberg also decreases engineering overhead. Instead of writing a compiler pass for constant propagation, we prompt the LLM to produce programs with constant loop bounds from the beginning. By encapsulating multiple tool-specific constraints within the initial prompt, Iceberg ensures that the generated programs seamlessly integrate with HLS tools, eliminating the need for extensive manual adjustments.

Beyond enforcing formatting constraints, we also modify the prompt to enhance diversity in the generated programs. Specifically, we instruct the LLM to make programs with varying characteristics, including a single or multiple functions, $Q$ loops, $M$ memory footprint, and computational workloads from a specified \textit{domain}, where $Q$, $M$, and \textit{domain} are randomly sampled parameters.

An interesting observation is that LLMs attempt to calculate memory footprints by summing array sizes, leading to the creation of programs with diverse scales. For the $domain$ parameter, we curate a set of computationally intensive domains from which the LLM selects, ensuring broad coverage of application types. Iceberg spans a wide range of domains, including scientific simulations, cryptography, signal processing, and medical imaging.

\subsection{Iterative generation}
A critical challenge when relying solely on the initial prompt is that LLMs often generate highly similar programs within certain application domains. For example, when prompted with the domain ``reinforcement learning," the model consistently produces the dynamic programming algorithm for Q-learning, limiting diversity in generated programs.

To address this issue, we introduce a simple yet effective technique. For each domain, we feed previously generated programs back into the prompt and apply zero-shot chain-of-thought (CoT)~\cite{cot} prompting, explicitly instructing the LLM to design a novel program that is significantly different from those already presented. This approach encourages the model to explore alternative algorithmic solutions within the same domain.

With this technique, program diversity improves substantially. For instance, instead of only generating Q-learning, the LLM can now produce programs resembling policy iteration, value iteration, PPO and other reinforcement learning algorithms. This ensures broader coverage of domain-specific variations, making the generated dataset more representative and useful for downstream tasks.
\section{Semi-synthetic training using Transformer Neural Process}
\label{sec:tnp}

\begin{figure*}[h]
    \centering
    \includegraphics[width=0.92\textwidth]{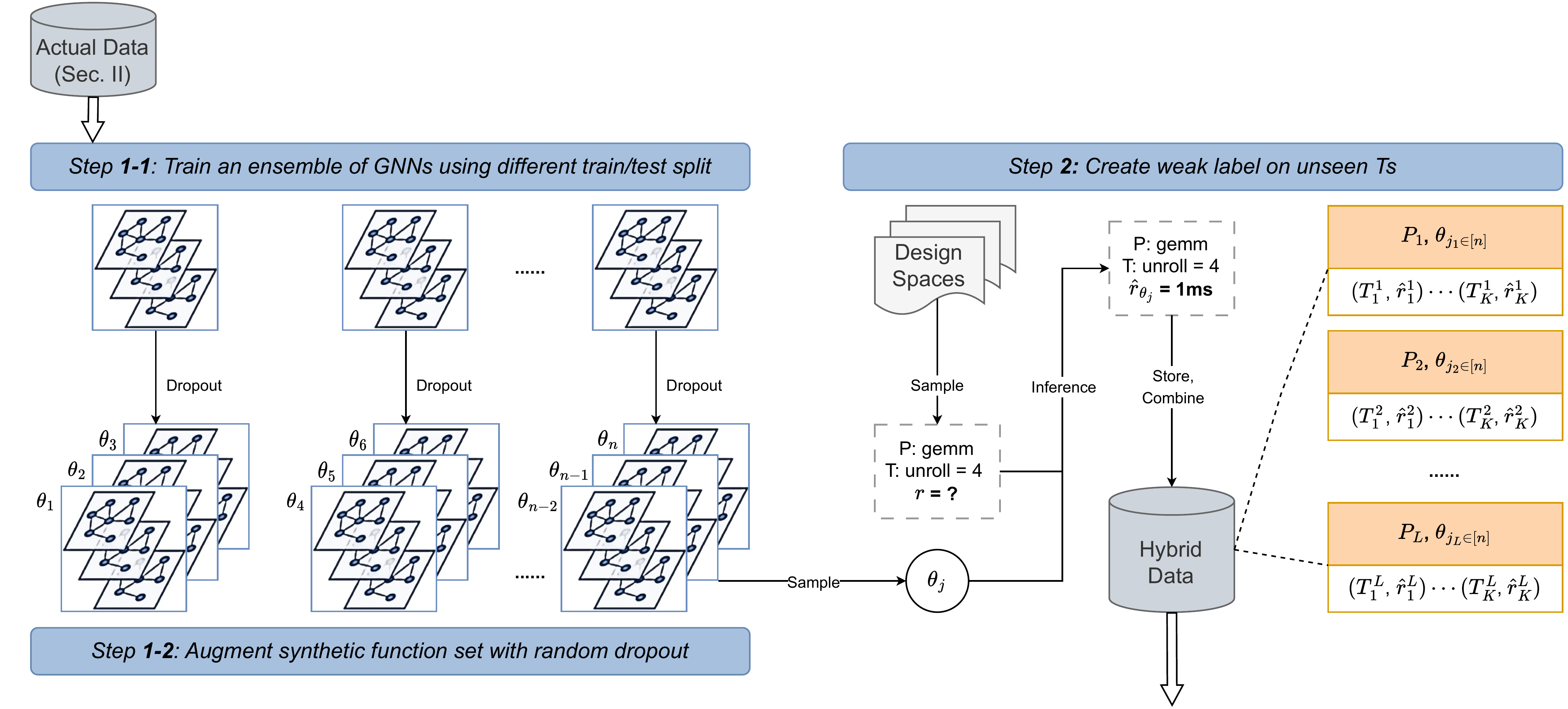}
    \caption{\small Creating weak labels. Given the actual training data and the design space of each program in the training set, we train an ensemble of GNNs and apply random dropout to each ensemble member to produce a set of synthetic functions, denoted as $\{\theta_j\}_{j\in[n]}$. We then randomly sample a training program $P$, a GNN parameterized by $\theta_{j}$, several unseen design configurations from the design space of $P$, and use the GNN to spawn weak labels for these configurations. We sample $L$ (Program, Function) pairs in total, denoted as $(P_i, \theta_{j_i})$, and sample $K$ design configurations $(T_1^i\cdots T_K^i)$ for each pair. Using the corresponding synthetic function, we predict a weak label of each configuration: $\hat{r}_{k\in[K]}^i\leftarrow \hat{r}_{\theta_{j_i}}(T_k^i)$. The resulting hybrid dataset combines data with weak labels and accurate labels.}
    \label{fig:weak_label}
    \vspace{-10pt}
\end{figure*}

While LLMs offer a promising approach to scale the P dimension, obtaining a sufficient number of design configurations (Ts) for each P remains a challenge in achieving robust modeling accuracy. To mitigate this issue, we first explain how in-context learning architectures, such as the Transformer Neural Process (TNP)~\cite{tnp}, facilitate learning from a hybrid dataset consisting of both actual and weak labels.

Building on this foundation, we then propose a novel method for spawning cost-effective yet ``related" weak labels to train the TNP model. With this approach, we improve the generalizability of the model while reducing the dependency on expensive evaluations.

\subsection{Background: Transformer Neural Process}
\begin{figure}[h]
    \centering
    \includegraphics[width=0.85\columnwidth]{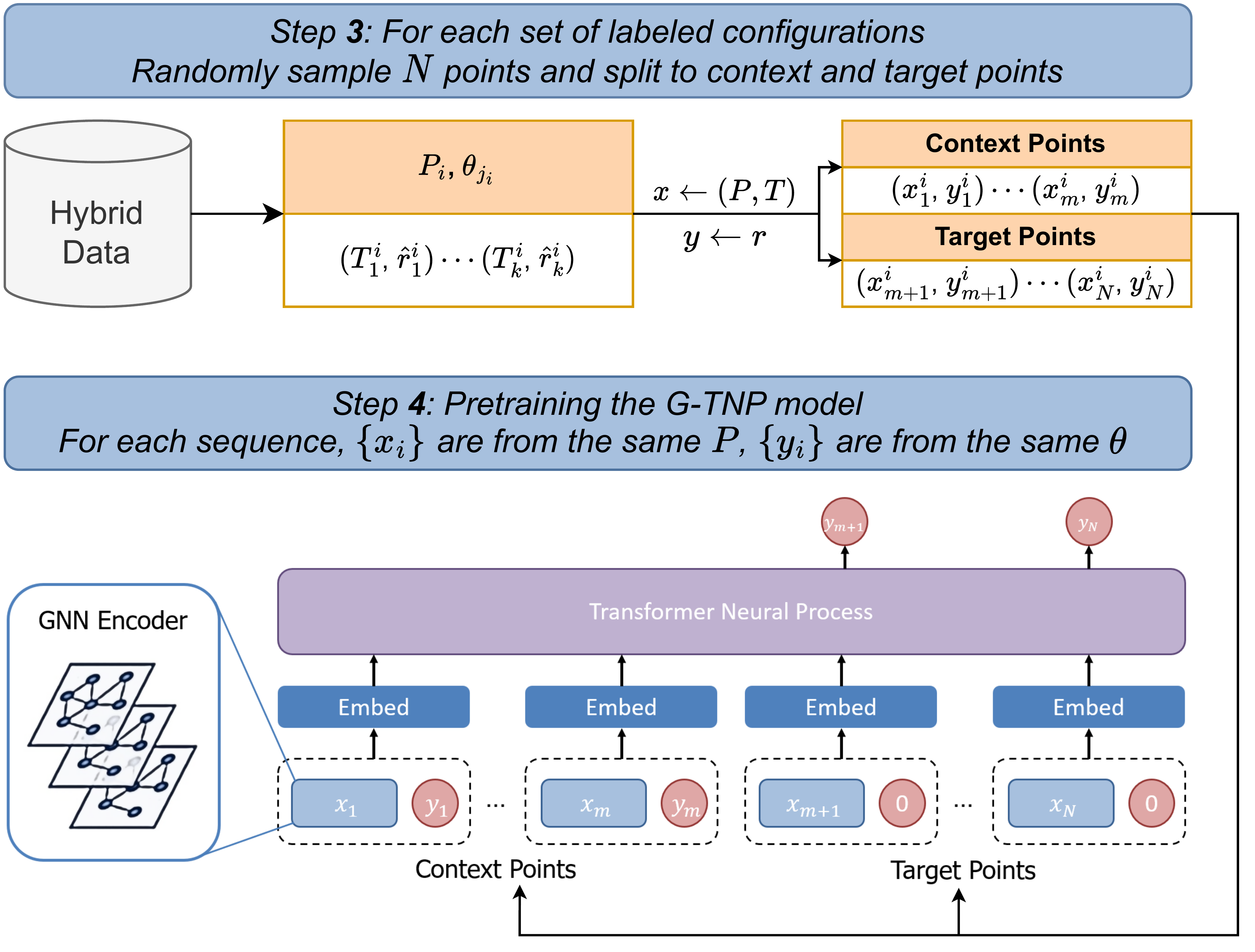}
    \caption{\small Pretraining the G-TNP model. From the hybrid dataset containing both actual and weak labels, we sample sequences $\{(x_i,y_i)\}$ and correspond to the same program $P_i$ and the same function $\theta_{j_i}$. This ensure that the $N$ points in each sequence share the same context. Then, we split the sequence to ``context points" and ``target points" and calculate the loss accordingly.}
    \label{fig:tnp_train}
    \vspace{-10pt}
\end{figure}

We first briefly explain how TNP could enable in-context meta-learning. Fig.~\ref{fig:tnp_train} illustrates the architecture of the diagonal Transformer Neural Process (TNP). During each training step, the model processes a sequence of input-output pairs:
$$\{(x_1,y_1),\cdots, (x_m, y_m),(x_{m+1}, y_{m+1}),\cdots,(x_N,y_N)\}$$

This sequence is divided into two parts: ``context points" and ``target points". The ``context points", $\{(x_1,y_1),\cdots, (x_m, y_m)\}$, capture the contextual information. For each context point, the dense vector $\bold{x_i}$ is concatenated with its corresponding scalar output $y_i$ to form a single token. This token is then encoded using a multi-layer perceptron (MLP).

Then, for the ``target points" $\{x_{m+1},\cdots,x_N\}$,  the goal is to predict the corresponding outputs, $\{y_{m+1},\cdots,y_N\}$ given the encoded contextual information. To achieve this, each target point $x_i$ is transformed into a token $(x_i, 0)$, where the output value is set to zero. This token is then processed through the Transformer encoder with the context tokens to compute its hidden embedding, which is subsequently decoded with an MLP to obtain the predicted output $y_i$.

Unlike traditional supervised learning, where the model learns a direct mapping from inputs to outputs, the TNP architecture leverages attention mechanisms to relate the target inputs $x_{m+1:N}$ to the contextual information $(x_{1:m},y_{1:m})$. This enables in-context adaptation at test time, allowing the model to generalize more effectively to unseen data. Multiple studies have demonstrated that in-context learning results in more robust performance than conventional supervised fine-tuning in few-shot learning settings~\cite{expt,lico}, making TNP a powerful approach for scenarios with limited labeled data.

In principle, TNP has two unique properties that distinguish it from pretrained LLMs: context invariance and target equivariance. These properties ensure that permuting either the context points or the target points does not affect the model’s predictions. This behavior is achieved by removing positional embeddings, allowing the model to treat input sequences as unordered sets rather than fixed sequences. Additionally, concatenating $x_i$ with $y_i$ explicitly captures the strong relationship between inputs and outputs, ensuring that the model learns dependencies independent of their ordering.

\subsection{Modeling HLS designs with TNP}
A key challenge in permitting in-context modeling for HLS designs is handling the variable input sizes of $x_i$. Specifically, for a program containing $K$ loops, the dimension of the corresponding design configuration becomes $3\cdot K$ as explained in Appx. \ref{appx:pragma_insertion}. This variability prohibits the direct application of the TNP architecture, which necessities $x_i$ to have a fixed dimension. To overcome this limitation, we observe that while $x_i$ must be a fixed-dimensional dense vector, it does not need to be the raw input. Instead, it can be a hidden embedding generated by an encoder, allowing variable-sized design configurations to be mapped into a consistent representation.

Building on the success of Graph Neural Networks (GNNs) in modeling HLS designs, we encode the embeddings using HARP’s encoder~\cite{harp}. This permits us to train a single TNP backbone with design configurations from different programs, overcoming the challenge of variable input sizes. We refer to this unique architecture as G-TNP.

For each sequence, our G-TNP model calculates the following loss:
$$\mathcal{L}(\phi_1, \phi_2)=\Pi_{i=m+1}^N p_{\phi_2}(y_i|h_{\phi_1}(x_i),h_{\phi_1}(x_{1:m}),y_{1:m})$$ where $p_{\phi_2}$ represents the predictive distribution modeled by the TNP with parameters $\phi_2$, and $h_{\phi_1}$ denotes the hidden representation generated by the GNN encoder with parameters $\phi_1$. This model architecture effectively leverages both graph-based structural information and in-context learning.

Despite enabling training across multiple programs, G-TNP can also model a distribution over functions. This is because it is intrinsically a Neural Process~\cite{garnelo2018neural} that learns a predictive distribution over target labels, conditioned on both context pairs and target inputs. This property allows us to perform data augmentation running cheap, synthetic functions, training G-TNP with both weak and accurate labels. More importantly, we can now augment the T dimension by introducing unseen Ts with weak labels from synthetic functions, rather than relying on computationally expensive HLS tools to generate them. However, the key challenge remains: how can we obtain cheap, yet useful, synthetic functions for HLS designs?

\subsection{Weak labels for HLS}

Several previous studies have explored the synergy between transformer-based architectures and synthetic functions. For instance, random Gaussian Processes (GPs) have been used as a source of synthetic functions in TNP. Similarly, in TabPFN~\cite{tabpfn}, a mixture of Bayesian Neural Networks (BNNs) and Structural Causal Models (SCMs) is employed to construct a prior for tabular data.

Our experiments with a naive source of synthetic functions yielded negative results (Appx.~\ref{appx:compare_syn}), which is unsurprising. Even with a perfect meta-learning model, the hypothesis class must be well-calibrated with the actual function and have low variance when this condition is met. An improper class can mislead the model, pulling it away from the true distribution.

To generate a class of synthetic functions that satisfy the necessary conditions, we draw inspiration from uncertainty estimation techniques in deep learning. Specifically, the ensemble method~\cite{lakshminarayanan2017simple}, naturally provides a set of well-calibrated functions closely related to the actual function. While training a large ensemble is computationally expensive, we further augment this set with MC-Dropout~\cite{gal2016dropout}, which creates diverse functions efficiently by simply sampling neuron masks, offering a cheaper yet effective alternative.

As shown in Fig. \ref{fig:weak_label}, we generate design configurations with weak labels for each program $P$ using function $\theta_j$ with the following mechanism:
$$\mathcal{S}_{P,\theta_j}=\{(T_k^P, \hat{r}_{\theta_j}(T_k^P))\}, k\in[K], T_k^P\sim \mathcal{D}_P, \theta_j\sim\mathcal{P}(\Theta|\mathcal{A})$$

Where $\theta_j$ is a GNN obtained by training on the the actual data $\mathcal{A}$ (Sec. \ref{sec:iceberg}) and applying random dropout. $\mathcal{D_P}$ denotes the design space of program $P$, and we sample $K$ design configurations $\{T_k^P\}_{k\in[K]}$ from $\mathcal{D_P}$. For each configuration, we obtain its weak label by running inference $\hat{r}_{\theta_j}(T_k^P)$. This procedure is repeated $L$ times to build a dataset of design configurations with weak labels. As shown in the orange boxes of Fig. \ref{fig:weak_label}, each distinct pair of $(P_i, \theta_{j_i})$ defines a unique context, and our hybrid dataset contains $L$ contexts in total.

To distinguish between the LLM-generated synthetic programs (Sec. \ref{sec:iceberg}) and our proposed weak label augmentation technique, we refer to the Iceberg dataset as the actual labels collected from the synthetic programs. In the following evaluation section, we explicitly indicate when weak labels are used for training the G-TNP model.
\section{Evaluation}

\subsection{Baselines}
For prediction, we compare Iceberg against two existing baselines: HARP~\cite{harp} and Hierarchical-MoE~\cite{li2025hierarchical}.

HARP employs a hierarchical graph architecture designed specifically for HLS, addressing the over-smoothing issue commonly observed in graph-based models. Hierarchical-MoE extends HARP by incorporating mixture-of-expert (MoE) layers at different levels of granularity, improving generalizability by capturing common structures.

\subsection{Benchmarks}
We assess the individual contributions of the G-TNP model architecture and the weak label generation by conducting evaluations under two different pretraining settings. This isolates the effects of the model architecture and the impact of synthetic data augmentation.

\textbf{Evaluating in-context modeling.} We follow the experimental setup of Hierarchical-MoE~\cite{li2025hierarchical}, the current state-of-the-art (SoTA) model for HLS in few-shot domain adaptation. We use the same training and testing split of the HLSyn dataset. When adapting to test programs, we select the same set of 50 design configurations through random sampling, utilizing them either as context points for in-context learning or as a fine-tuning dataset for supervised adaptation.

\textbf{Pretraining with the Iceberg dataset.} We train all baselines on Iceberg’s dataset and compare the pretrained models on two test sets: (1) the HLSyn dataset, a public benchmark for HLS modeling, and (2) six challenging real-world applications sourced from the Rosetta benchmark\cite{rosetta} and the attention layer\cite{attention}.

Unlike the experimental setup in Hierarchical-MoE, we collect labeled design configurations running Vitis HLS 2023.2~\cite{vitis23}, a more recent toolchain. Our evaluation includes 16 test programs in total --- 10 from HLSyn and 6 real-world applications (Table \ref{tab:benchmark_descriptions}) --- forming a more comprehensive benchmark.

\begin{table}[h]
    \centering
    \caption{Real-world benchmark descriptions}
    \label{tab:benchmark_descriptions}
    \begin{tabular}{l l}
        \toprule
        Benchmark & Description \\
        \midrule
        att-3mm & First three matrix multiplications of the attention layer \\
        att-3mm-fuse & Changing the program structure of att-3mm \\
        3d-rendering & Video processing \\
        optical-flow & Video processing \\
        conv2d & Single 2D convolution layer \\
        knn & Machine learning \\
        \bottomrule
    \end{tabular}
\end{table}

\subsection{Training details}
For prediction, we test four variations to analyze the impact of weak labels and fine-tuning: (1) Ice-A, where the G-TNP model is trained using only actual labels; (2) Ice-H, where the model is trained on a hybrid dataset that includes both actual and weak labels; (3) Ice-A-FT, where Ice-A is fine-tuned on few-shot adaptation data; and (4) Ice-H-FT, where Ice-H is fine-tuned on few-shot adaptation data. Recent studies have shown that fine-tuning in-context models on test data improves performance~\cite{mixtabpfn}.

All models are trained with a mean squared error (MSE) loss. For Ice-H, the ratio of weak to actual label is set to 0.5. During fine-tuning, we use only the few-shot adaptation data and fine-tune for 200 steps.

\subsection{Optimization details}
It is well known that improvements in modeling accuracy do not always translate to better optimization results. To assess model performance in an optimization setting, we develop an offline evaluation protocol. Specifically, we sample a set of designs and get their actual HLS labels. Each model then predicts the performance of these samples, and we select the top $K$ designs based on the model's predictions. Using the actual labels, we assess the best-performing design among the top-$K$ selections, defining this metric as best@$K$.

Following prior work, we set a resource utilization boundary at 0.8. Since Iceberg focuses on performance modeling, we first run the resource and classification models from Hierarchical-MoE to filter out valid designs. We then apply the Iceberg performance model to pick the top-$K$ designs.

\subsection{Hyperparameters and Dataset statistics}

Table \ref{tab:hyperparams} lists the hyperparameters of the Transformer encoder employed in TNP, along with the training hyperparameters. Since the Transformer encoder is relatively small ($\sim$591K parameters), our method introduces minimal overhead at inference time.

\begin{table}[h]
    \centering
    \caption{Transformer encoder hyperparameters}
    \label{tab:hyperparams}
    \begin{tabular}{l c}
        \toprule
        Hyperparameter & Value \\
        \midrule
        Number of layers & 6 \\
        Hidden dimension & 128 \\
        Number of heads & 8 \\
        Feedforward dimension & 128 \\
        Dropout rate & 0.1 \\
        Learning rate & 0.0005 \\
        Number of steps (HLSyn) & 20000 \\
        Number of steps (Iceberg) & 64000 \\
        Optimizer & AdamW \\
        $\beta_1$ & 0.9 \\
        $\beta_2$ & 0.99 \\
        Weight decay & $1 \times 10^{-5}$ \\
        \bottomrule
    \end{tabular}
    \vspace{-10pt}
\end{table}

Table~\ref{tab:ice_stats} summarizes statistics from the Iceberg dataset. For pretraining the Ice-H baselines, we consistently sample $L=300$ $(P, \theta)$ pairs,  each associated with $K = 100$ design configurations, as defined in Fig. \ref{fig:weak_label}. Note that, in this study, training utilizes only the $214$ programs that have actual labels, due to the limited zero-shot accuracy of existing models on unseen programs. However, all $3000+$ synthesizable programs are publicly released to support future research.

\begin{table}[h]
    \centering
    \caption{Iceberg dataset statistics}
    \label{tab:ice_stats}
    \begin{tabular}{l c}
        \toprule
        Statistic & Value \\
        \midrule
        Total programs generated & 4449\\
        Synthesizable programs & 3401 \\
        Programs with actual labels & 214 \\
        Total actual label & 14840 \\
        \bottomrule
    \end{tabular}
\end{table}

Our approach to weak label generation is motivated by the observation that existing GNN models demonstrate acceptable accuracy on unseen configurations within training programs (see Table~\ref{tab:weak_label_acc}). However, we find a significant drop in accuracy when these GNNs are trained on the Iceberg dataset. This decline occurs because the Iceberg dataset includes programs that are significantly more diverse in structure and memory footprint, thus making accurate predictions considerably more challenging. This reduced accuracy in weak label generation might explain why the difference in test loss between Ice-A and Ice-H becomes less pronounced when pretrained on the Iceberg dataset (Appx. Fig. \ref{fig:iceberg_compare}). Further investigation is left to future work.

\begin{table}[h]
    \centering
    \caption{\small Weak label accuracy. Each GNN in the ensemble is trained with a random dataset split (15\% validation, 15\% test). We trained 19 GNNs on HLSyn (Table \ref{tab:res_hlsyn}) and 14 on Iceberg (Figs. \ref{fig:iceberg_res_1}, \ref{fig:iceberg_res_2}). Mean and standard deviation of test MSE across the ensemble are reported.}
    \label{tab:weak_label_acc}
    \begin{tabular}{l c c}
    \toprule
    & On HLSyn & On Iceberg\\
    \midrule
    Average test MSE & 0.047  & 0.321\\
    Std         & 0.005  & 0.117\\
    \bottomrule
    \end{tabular}
\end{table}

\subsection{Results: the Iceberg dataset improves generalization}
To demonstrate that the Iceberg dataset enhances adaptation to downstream tasks, we designate HLSyn as the target dataset and evaluate two baseline models, HARP and Hierarchical-MoE (H-MoE). We train each model either from scratch on HLSyn or with pretraining on Iceberg followed by fine-tuning on HLSyn. With pretraining, HARP’s testing accuracy risises by 80\%, while H-MoE’s accuracy increases by 10\%. These results indicate that synthetic programs with actual labels enhance model generalization, regardless of the model architecture. Detailed results are provided in Appx. \ref{appx:details}.

\subsection{Results: In-context modeling on HLSyn}
\label{sec:eval_hlsyn}

\begin{table*}[h]
    \centering
    \caption{\small Comparison of the Iceberg performance model with two baselines on the HLSyn dataset, where the programs are split into training and testing sets. The baselines include HARP (fine-tuned GNN with hierarchical graph) and H-MoE (fine-tuned Hierarchical-MoE). We evaluate Ice-A (G-TNP trained with actual labels), Ice-H (G-TNP trained with both actual and weak labels), Ice-A-FT (fine-tuned Ice-A), and Ice-H-FT (fine-tuned Ice-H). Each experiment is repeated three times with random seeds 1, 2, and 3.}
    \begin{tabular}{c|ccccccc}
    \hline
             & fdtd-2d-large & gemver-medium & syr2k & gemm-p & jacobi-2d & trmm-opt & geomean\\
    \hline
   HARP         & 0.35 (0.15) & 0.03 (0.02) & 0.03 (0.02) & 0.18 (0.06) & 0.15 (0.04) & 0.30 (0.20) & 0.12 (0.05)\\
   H-MoE        & 0.26 (0.10) & 0.02 (0.01) & 0.02 (0.02) & 0.12 (0.01) & 0.12 (0.02) & 0.18 (0.10) & 0.08 (0.03)\\
   Ice-A    & 1.40 (0.18) & 0.06 (0.02) & 0.03 (0.01) & 0.59 (0.12) & 1.27 (0.07) & 0.47 (0.04) & 0.31 (0.04) \\
   Ice-H    & 0.95 (0.22) & 0.05 (0.01) & 0.02 (0.01) & 0.25 (0.09) & 0.08 (0.01) & 0.36 (0.08) & 0.14 (0.03)\\
   Ice-A-FT & 0.82 (0.65) & 0.00 (0.00) & 0.02 (0.00) & 0.30 (0.13) & 0.02 (0.01) & 0.15 (0.06) & \textbf{0.06} (0.02) \\
   Ice-H-FT & 0.24 (0.40) & 0.01 (0.00) & 0.04 (0.02) & 0.12 (0.03) & 0.02 (0.00) & 0.17 (0.07) & \textbf{0.06} (0.03) \\
    \hline
    \end{tabular}
    \vspace{-5pt}
    \label{tab:res_hlsyn}
\end{table*}

\textbf{In-context prediction.} Table \ref{tab:res_hlsyn} summarizes the prediction accuracy on the dataset released by H-MoE. Each baseline is tested using three different random seeds (1,2,3), and we report the mean and standard deviation of the MSE loss. 

The Ice-A baseline suffers from severe overfitting, performing $1.72\times$ worse than H-MoE. This is primarily due to the small training dataset, which consists of only 33 training programs with a few hundred designs per program. The GNN encoder struggles to generalize when the structure or features of the testing programs differ significantly from the training programs. Additionally, the Transformer backbone, which has more parameters than the GNN encoder, is prone to overfitting in this low-data setting.

However, Ice-A-FT significantly boosts performance, surpassing H-MoE by 25\%, demonstrating that fine-tuning at test time can substantially boost accuracy, even for in-context models. With Ice-A-FT achieving superior performance over H-MoE, we conclude that our in-context model which leverages the unique combination of a GNN encoder and the TNP architecture attains better accuracy when adapting to unseen programs.

\textbf{Augmenting weak labels.} With 50\% weak labels, covering a wider range of the design space for each training program, Ice-H achieves 55\% lower MSE compared to Ice-A. This demonstrates that training with hybrid data effectively mitigates overfitting, particularly when the amount of training data is limited. In Appx. \ref{appx:scale_f}, we show that the best modeling accuracy cannot be achieved when the weak labels come from a single synthetic function.

While weak label augmentation significantly enhances modeling accuracy, both Ice-A and Ice-H fall short of outperforming the H-MoE baseline. This suggests that in-context models struggle with adaptability when the training dataset is insufficient.

\subsection{Results: Pretraining G-TNP on the Iceberg dataset}
\label{sec:eval_iceberg}

\begin{figure}[h]
    \centering
    \vspace{-5pt}
    \subfloat[Adapting to 10 programs from HLSyn]{
        \includegraphics[width=0.51\columnwidth]{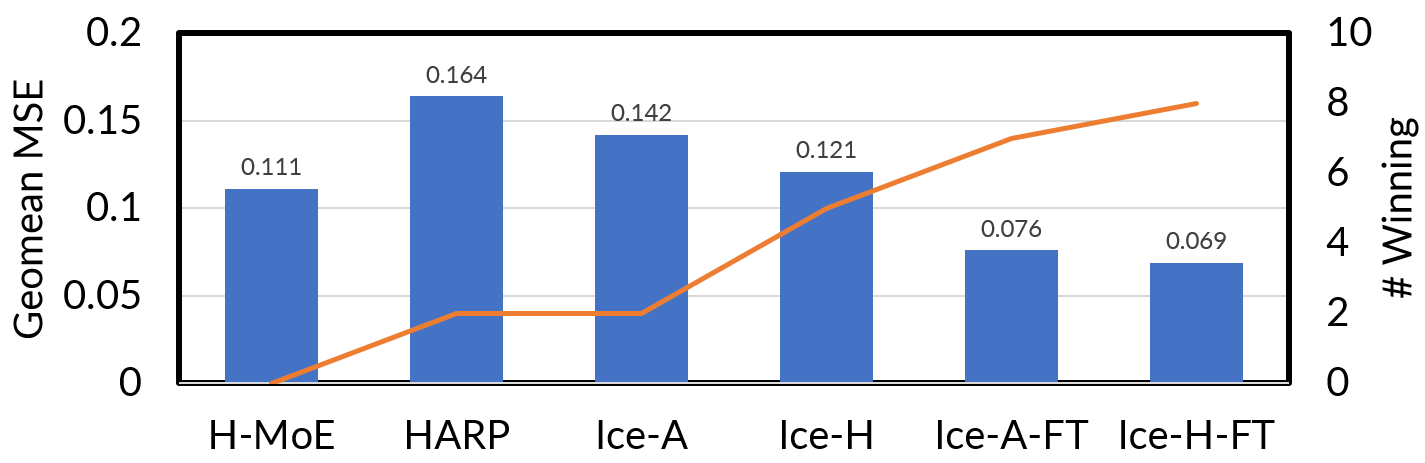}
        \label{fig:iceberg_res_1}
    }
    \hspace{1pt}
    \subfloat[Adapting to 6 real world programs]{
        \includegraphics[width=0.40\columnwidth]{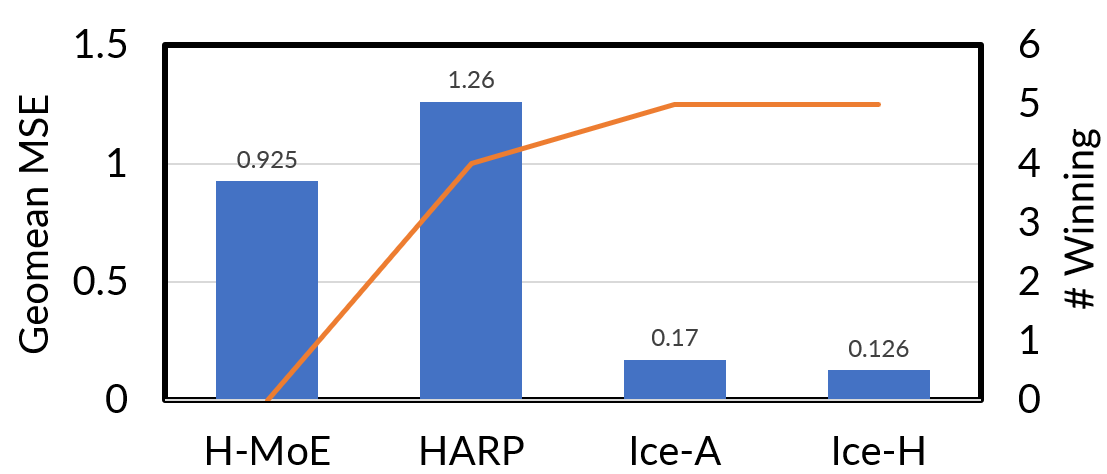}
        \label{fig:iceberg_res_2}
    }
    \caption{\small Comparison of the Iceberg performance model with other baselines. Pretrained on the Iceberg dataset (with or without weak labels) and adapted to two different test sets.}
    \label{fig:editing}
\end{figure}

In the previous set of experiments, we observed that due to the limited diversity in the training set, in-context models --- even when trained with weak labels from unseen design configurations --- could not surpass existing supervised learning baselines. However, as shown in Fig.~\ref{fig:iceberg_res_1}, this limitation is alleviated when we switch the training dataset (for all baselines, including HARP and H-MoE) to the Iceberg dataset.

When testing on HLSyn programs, we observe that the gap between in-context prediction models and prior SoTA models is narrowing. Notably, both Ice-A and Ice-H provide better accuracy than HARP, despite relying solely on in-context adaptation at test time, without any parameter updates. This confirms that a high-quality synthetic pretraining dataset can significantly enhance the performance of in-context models.

Our findings on fine-tuning follow the same trend. The geometric mean MSE of Ice-A-FT across 10 test programs rises by 46.5\% over Ice-A, while Ice-H-FT improves by 43.0\% over Ice-H. Additionally, the fine-tuned models outperform H-MoE by 31.5\% (Ice-A-FT) and 37.9\% (Ice-H-FT), further establishing the effectiveness of in-context learning combined with fine-tuning.

We observe an even more significant improvement of in-context models over SoTA supervised learning models when tested on the six real-world applications. As illustrated in Fig. \ref{fig:iceberg_res_2}, even without fine-tuning, Ice-H surpasses H-MoE’s geometric mean MSE by 86.4\%. This further validates both the effectiveness of our synthetic data augmentation method and the robustness of in-context prediction models in adapting to real-world HLS designs.

Although we collected a large number of actual performance labels by running AutoDSE on synthetic programs, we still observe slight improvements when augmenting with weak labels for unseen design configurations. Ice-H consistently beats Ice-A on both test sets, despite the only difference being that Ice-A is trained solely on actual labels from HLS tools, while Ice-H incorporates a mix of actual and weak labels.

While scaling high-fidelity actual labels is ideal, a diverse set of weak labels remains beneficial when the available actual labels are insufficient to mitigate distribution shift.

\subsection{Results: Forward optimization}
\begin{figure}
    \centering
    \vspace{-10pt}
    \subfloat[Adapting to HLSyn]{
        \includegraphics[width=0.45\columnwidth]{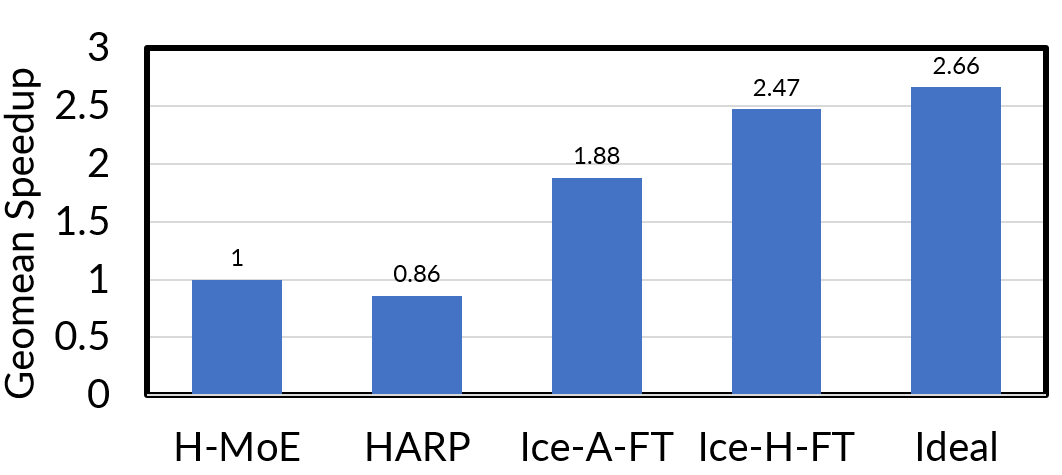}
        \label{fig:best_1_hlsyn}
    }
    \hspace{2pt}
    \subfloat[Adapting to real world programs]{
        \includegraphics[width=0.45\columnwidth]{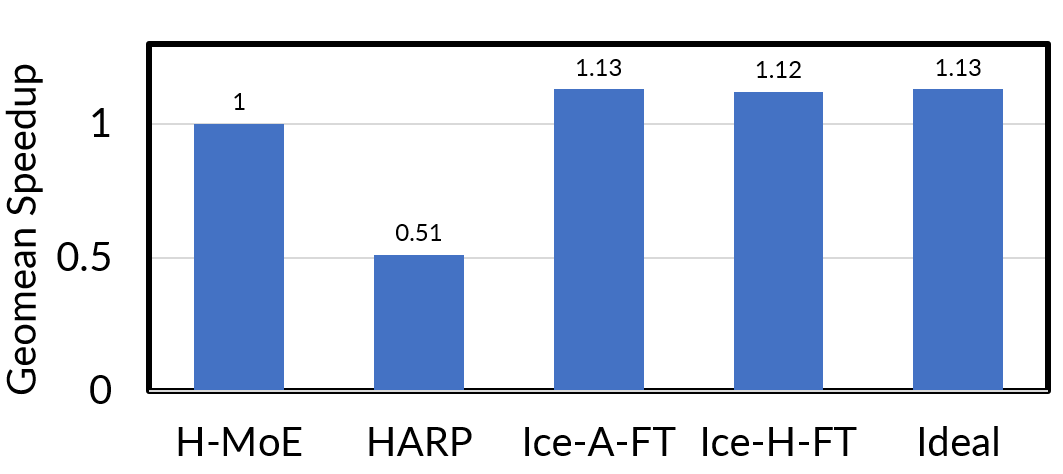}
        \label{fig:best_1_real}
    }
    \caption{\small Offline design optimization: best@1}
    \label{fig:opt}
    \vspace{-10pt}
\end{figure}

Fig.~\ref{fig:opt} presents the results of our model-based forward optimization experiment, where we evaluate performance using the best@1 metric. If all model-selected designs are invalid, we substitute the result with the performance of the default design point. All baselines are pretrained on Iceberg and adapted to either the HLSyn dataset or the six real-world applications before performing optimization. This evaluation measures how well each model identifies the best-performing design configuration in an offline setting.

We observe that better modeling accuracy generally translates to better optimization performance. Specifically, Ice-H-FT achieves the highest best@1 metric when adapting to the HLSyn dataset. Additionally, when compared to the actual top design within the explored samples (denoted as ``Ideal"), the performance gap remains small, indicating that a strong prediction model is not only effective at accurately estimating the majority of design performances but also at identifying the highest-performing configurations.

However, we observe that improvements in modeling accuracy do not always translate directly into better design optimization performance. Despite the substantial gains in prediction accuracy, the Iceberg baselines (Ice-A-FT, Ice-H-FT) do not outperform H-MoE by a large margin on real-world applications. This suggests that while better modeling improves prediction quality, its impact on optimization may be limited in certain scenarios. We leave online evaluation as future work to further investigate the generalizability of the models.

\section{Acknowledgement}
This work was partially supported by NSF grants 2211557, 1937599,  2119643, and 2303037, NSF CAREER Grant 2341040, SRC JUMP 2.0 PRISM Center, NASA, Okawa Foundation, Amazon Research, Cisco, Picsart, Snapchat, and the CDSC industrial partners. The authors would also like to thank AMD/Xilinx for HACC equipment donation, Marci Baun for editing the paper, and Hanyu Wang and Su Zheng for their feedback. J. Cong has a financial interest in AMD.

\section{Conclusion}
In this work, we explore synthetic data for improving HLS modeling. We introduce Iceberg, a set of LLM-generated synthetic programs for HLS tools, and generate weak labels for unseen design configurations using an ensemble of GNNs. To learn from both actual and synthetic data, we leverage G-TNP, an in-context prediction model integrating a GNN encoder and a Transformer Neural Process. We achieve significant accuracy improvements over state-of-the-art HLS prediction methods.

\clearpage
\bibliographystyle{IEEEtran}
\bibliography{main}

\begin{thebibliography}{10}
\providecommand{\url}[1]{#1}
\csname url@samestyle\endcsname
\providecommand{\newblock}{\relax}
\providecommand{\bibinfo}[2]{#2}
\providecommand{\BIBentrySTDinterwordspacing}{\spaceskip=0pt\relax}
\providecommand{\BIBentryALTinterwordstretchfactor}{4}
\providecommand{\BIBentryALTinterwordspacing}{\spaceskip=\fontdimen2\font plus
\BIBentryALTinterwordstretchfactor\fontdimen3\font minus \fontdimen4\font\relax}
\providecommand{\BIBforeignlanguage}[2]{{%
\expandafter\ifx\csname l@#1\endcsname\relax
\typeout{** WARNING: IEEEtran.bst: No hyphenation pattern has been}%
\typeout{** loaded for the language `#1'. Using the pattern for}%
\typeout{** the default language instead.}%
\else
\language=\csname l@#1\endcsname
\fi
#2}}
\providecommand{\BIBdecl}{\relax}
\BIBdecl

\bibitem{cong2011high}
J.~Cong, B.~Liu, S.~Neuendorffer, J.~Noguera, K.~Vissers, and Z.~Zhang, ``High-level synthesis for {FPGA}s: From prototyping to deployment,'' \emph{IEEE Transactions on Computer-Aided Design of Integrated Circuits and Systems}, vol.~30, no.~4, pp. 473--491, 2011.

\bibitem{cong2022fpga}
J.~Cong, J.~Lau, G.~Liu, S.~Neuendorffer, P.~Pan, K.~Vissers, and Z.~Zhang, ``{FPGA} {HLS} today: successes, challenges, and opportunities,'' \emph{ACM Transactions on Reconfigurable Technology and Systems (TRETS)}, vol.~15, no.~4, pp. 1--42, 2022.

\bibitem{verilogeval}
\BIBentryALTinterwordspacing
M.~Liu, N.~Pinckney, B.~Khailany, and H.~Ren, ``{VerilogEval}: Evaluating large language models for {V}erilog code generation,'' 2023. [Online]. Available: \url{https://arxiv.org/abs/2309.07544}
\BIBentrySTDinterwordspacing

\bibitem{zhang2024mg}
Y.~Zhang, Z.~Yu, Y.~Fu, C.~Wan, and Y.~C. Lin, ``{MG}-{V}erilog: Multi-grained dataset towards enhanced {LLM}-assisted {V}erilog generation,'' in \emph{2024 IEEE LLM Aided Design Workshop (LAD)}.\hskip 1em plus 0.5em minus 0.4em\relax IEEE, 2024, pp. 1--5.

\bibitem{cui2024origen}
F.~Cui, C.~Yin, K.~Zhou, Y.~Xiao, G.~Sun, Q.~Xu, Q.~Guo, D.~Song, D.~Lin, X.~Zhang \emph{et~al.}, ``Origen: Enhancing {RTL} code generation with code-to-code augmentation and self-reflection,'' \emph{arXiv preprint arXiv:2407.16237}, 2024.

\bibitem{scalehls}
H.~Ye, C.~Hao, J.~Cheng, H.~Jeong, J.~Huang, S.~Neuendorffer, and D.~Chen, ``{ScaleHLS}: A new scalable high-level synthesis framework on multi-level intermediate representation,'' in \emph{2022 IEEE Iternational Symposium on High-Performance Computer Architecture (HPCA)}.\hskip 1em plus 0.5em minus 0.4em\relax IEEE, 2022, pp. 741--755.

\bibitem{streamhls}
S.~Basalama and J.~Cong, ``Stream-{HLS}: Towards automatic dataflow acceleration,'' \emph{arXiv e-prints}, pp. arXiv--2501, 2025.

\bibitem{lightningsimv2}
R.~Sarkar, R.~Paul, and C.~C. Hao, ``{LightningSimV2}: Faster and scalable simulation for high-level synthesis via graph compilation and optimization,'' in \emph{2024 IEEE 32nd Annual International Symposium on Field-Programmable Custom Computing Machines (FCCM)}.\hskip 1em plus 0.5em minus 0.4em\relax IEEE, 2024, pp. 104--114.

\bibitem{sisyphus}
S.~Pouget, L.-N. Pouchet, and J.~Cong, ``A unified framework for automated code transformation and pragma insertion,'' in \emph{Proceedings of the 2025 ACM/SIGDA International Symposium on Field Programmable Gate Arrays}, 2025, pp. 187--198.

\bibitem{allo}
H.~Chen, N.~Zhang, S.~Xiang, Z.~Zeng, M.~Dai, and Z.~Zhang, ``{A}llo: {A} programming model for composable accelerator design,'' \emph{{P}roceedings of the {ACM} on {P}rogramming {L}anguages}, vol.~8, no. {PLDI}, pp. 593--620, 2024.

\bibitem{hlsdebug}
L.~J. Wan, H.~Ye, J.~Wang, M.~Jha, and D.~Chen, ``An iteratively-refined dataset for high-level synthesis functional verification through {LLM}-aided bug injection,'' in \emph{2024 IEEE LLM Aided Design Workshop (LAD)}.\hskip 1em plus 0.5em minus 0.4em\relax IEEE, 2024, pp. 1--6.

\bibitem{hlspilot}
\BIBentryALTinterwordspacing
C.~Xiong, C.~Liu, H.~Li, and X.~Li, ``{HLSPilot}: {LLM}-based high-level synthesis,'' 2024. [Online]. Available: \url{https://arxiv.org/abs/2408.06810}
\BIBentrySTDinterwordspacing

\bibitem{collini2024c2hlsc}
L.~Collini, S.~Garg, and R.~Karri, ``{C2HLSC}: Can {LLMs} bridge the software-to-hardware design gap?'' \emph{arXiv preprint arXiv:2406.09233}, 2024.

\bibitem{ralad}
H.~Xu, H.~Hu, and S.~Huang, ``Optimizing high-level synthesis designs with retrieval-augmented large language models,'' in \emph{2024 IEEE LLM Aided Design Workshop (LAD)}.\hskip 1em plus 0.5em minus 0.4em\relax IEEE, 2024, pp. 1--5.

\bibitem{hlsrepair}
K.~Xu, G.~L. Zhang, X.~Yin, C.~Zhuo, U.~Schlichtmann, and B.~Li, ``Automated {C}/{C}++ {P}rogram {R}epair for {H}igh-{L}evel {s}ynthesis via {L}arge {L}anguage {M}odels,'' in \emph{Proceedings of the 2024 ACM/IEEE International Symposium on Machine Learning for CAD}, 2024, pp. 1--9.

\bibitem{llmaccel}
C.~Hong, S.~Bhatia, A.~Haan, S.~K. Dong, D.~Nikiforov, A.~Cheung, and Y.~S. Shao, ``{LLM}-aided compilation for tensor accelerators,'' in \emph{2024 IEEE LLM Aided Design Workshop (LAD)}.\hskip 1em plus 0.5em minus 0.4em\relax IEEE, 2024, pp. 1--14.

\bibitem{gpt4aigchip}
Y.~Fu, Y.~Zhang, Z.~Yu, S.~Li, Z.~Ye, C.~Li, C.~Wan, and Y.~C. Lin, ``{GPT}4{AIG}chip: {T}owards next-generation {AI} accelerator design automation via large language models,'' in \emph{2023 {IEEE/ACM} {I}nternational {C}onference on {C}omputer {A}ided {D}esign ({ICCAD})}.\hskip 1em plus 0.5em minus 0.4em\relax IEEE, 2023, pp. 1--9.

\bibitem{ironmanpro}
N.~Wu, Y.~Xie, and C.~Hao, ``{IRONMAN-PRO}: Multiobjective design space exploration in {HLS} via reinforcement learning and graph neural network-based modeling,'' \emph{IEEE Transactions on Computer-Aided Design of Integrated Circuits and Systems}, vol.~42, no.~3, pp. 900--913, 2022.

\bibitem{harp}
A.~Sohrabizadeh, Y.~Bai, Y.~Sun, and J.~Cong, ``Robust {GNN}-based representation learning for {HLS},'' in \emph{2023 IEEE/ACM International Conference on Computer Aided Design (ICCAD)}, 2023, pp. 1--9.

\bibitem{murphy2024balor}
E.~Murphy and L.~Josipovi{\'c}, ``Balor: {HLS} source code evaluator based on custom graphs and hierarchical {GNN}s,'' in \emph{Intl. Conference on Computer-Aided Design (ICCAD)}, 2024, pp. 1--9.

\bibitem{sourcetopostroute}
M.~Gao, J.~Zhao, Z.~Lin, and M.~Guo, ``Hierarchical source-to-post-route {QoR} prediction in high-level synthesis with {GNNs},'' in \emph{2024 Design, Automation \& Test in Europe Conference \& Exhibition (DATE)}.\hskip 1em plus 0.5em minus 0.4em\relax IEEE, 2024, pp. 1--6.

\bibitem{bai2024learning}
Y.~Bai, A.~Sohrabizadeh, Z.~Ding, R.~Liang, W.~Li, D.~Wang, H.~Ren, Y.~Sun, and J.~Cong, ``Learning to compare hardware designs for high-level synthesis,'' in \emph{Proceedings of the 2024 ACM/IEEE International Symposium on Machine Learning for CAD}, 2024, pp. 1--7.

\bibitem{powergear}
Z.~Lin, Z.~Yuan, J.~Zhao, W.~Zhang, H.~Wang, and Y.~Tian, ``{PowerGear}: Early-stage power estimation in {FPGA} {HLS} via heterogeneous edge-centric {GNNs},'' in \emph{2022 Design, Automation \& Test in Europe Conference \& Exhibition (DATE)}.\hskip 1em plus 0.5em minus 0.4em\relax IEEE, 2022, pp. 1341--1346.

\bibitem{qin2024cross}
Z.~Qin, Y.~Bai, A.~Sohrabizadeh, Z.~Ding, Z.~Hu, Y.~Sun, and J.~Cong, ``Cross-modality program representation learning for electronic design automation with high-level synthesis,'' in \emph{Proceedings of the 2024 ACM/IEEE International Symposium on Machine Learning for CAD}, 2024, pp. 1--12.

\bibitem{li2025hierarchical}
W.~Li, D.~Wang, Z.~Ding, A.~Sohrabizadeh, Z.~Qin, J.~Cong, and Y.~Sun, ``{H}ierarchical mixture of experts: {G}eneralizable learning for {h}igh-{l}evel {s}ynthesis,'' in \emph{Proceedings of the AAAI Conference on Artificial Intelligence}, vol.~39, no.~17, 2025, pp. 18\,476--18\,484.

\bibitem{active-cem}
Z.~Ding, A.~Sohrabizadeh, W.~Li, Z.~Qin, Y.~Sun, and J.~Cong, ``Efficient task transfer for {HLS} {DSE},'' in \emph{Proceedings of the 43rd IEEE/ACM International Conference on Computer-Aided Design}, 2024, pp. 1--9.

\bibitem{hlsyn}
Y.~Bai, A.~Sohrabizadeh, Z.~Qin, Z.~Hu, Y.~Sun, and J.~Cong, ``Towards a comprehensive benchmark for high-level synthesis targeted to {FPGA}s,'' \emph{Advances in Neural Information Processing Systems}, vol.~36, pp. 45\,288--45\,299, 2023.

\bibitem{hlsfactory}
S.~Abi-Karam, R.~Sarkar, A.~Seigler, S.~Lowe, Z.~Wei, H.~Chen, N.~Rao, L.~John, A.~Arora, and C.~Hao, ``{HLSFactory}: A framework empowering high-level synthesis datasets for machine learning and beyond,'' in \emph{Proceedings of the 2024 ACM/IEEE International Symposium on Machine Learning for CAD}, 2024, pp. 1--9.

\bibitem{merlin}
J.~Cong, M.~Huang, P.~Pan, Y.~Wang, and P.~Zhang, ``Source-to-source optimization for {HLS},'' \emph{{FPGA}s for Software Programmers}, pp. 137--163, 2016.

\bibitem{ye2024hida}
H.~Ye, H.~Jun, and D.~Chen, ``{HIDA}: A hierarchical dataflow compiler for high-level synthesis,'' in \emph{Proceedings of the 29th ACM International Conference on Architectural Support for Programming Languages and Operating Systems, Volume 1}, 2024, pp. 215--230.

\bibitem{autodse}
\BIBentryALTinterwordspacing
A.~Sohrabizadeh, C.~H. Yu, M.~Gao, and J.~Cong, ``{AutoDSE}: Enabling software programmers to design efficient {FPGA} accelerators,'' \emph{ACM Trans. Des. Autom. Electron. Syst.}, vol.~27, no.~4, Feb. 2022. [Online]. Available: \url{https://doi.org/10.1145/3494534}
\BIBentrySTDinterwordspacing

\bibitem{tnp}
T.~Nguyen and A.~Grover, ``Transformer {N}eural {P}rocesses: {U}ncertainty-aware meta learning via sequence modeling,'' \emph{arXiv preprint arXiv:2207.04179}, 2022.

\bibitem{expt}
T.~Nguyen, S.~Agrawal, and A.~Grover, ``Expt: Synthetic pretraining for few-shot experimental design,'' \emph{Advances in Neural Information Processing Systems}, vol.~36, pp. 45\,856--45\,869, 2023.

\bibitem{tabpfn}
\BIBentryALTinterwordspacing
N.~Hollmann, S.~Müller, K.~Eggensperger, and F.~Hutter, ``{TabPFN}: A transformer that solves small tabular classification problems in a second,'' 2023. [Online]. Available: \url{https://arxiv.org/abs/2207.01848}
\BIBentrySTDinterwordspacing

\bibitem{rosetta}
Y.~Zhou, U.~Gupta, S.~Dai, R.~Zhao, N.~Srivastava, H.~Jin, J.~Featherston, Y.-H. Lai, G.~Liu, G.~A. Velasquez \emph{et~al.}, ``Rosetta: A realistic high-level synthesis benchmark suite for software programmable {FPGA}s,'' in \emph{Proceedings of the 2018 ACM/SIGDA International Symposium on Field-Programmable Gate Arrays}, 2018, pp. 269--278.

\bibitem{attention}
A.~Vaswani, N.~Shazeer, N.~Parmar, J.~Uszkoreit, L.~Jones, A.~N. Gomez, {\L}.~Kaiser, and I.~Polosukhin, ``Attention is all you need,'' \emph{Advances in Neural Information Processing Systems}, vol.~30, 2017.

\bibitem{vitis23}
AMD/Xilinx, ``{Vitis HLS 2023.2},'' \url{https://docs.amd.com/r/en-US/ug1399-vitis-hls/Introduction}, 2023.

\bibitem{cot}
\BIBentryALTinterwordspacing
J.~Wei, X.~Wang, D.~Schuurmans, M.~Bosma, B.~Ichter, F.~Xia, E.~Chi, Q.~Le, and D.~Zhou, ``{Chain-of-Thought} prompting elicits reasoning in large language models,'' 2023. [Online]. Available: \url{https://arxiv.org/abs/2201.11903}
\BIBentrySTDinterwordspacing

\bibitem{lico}
T.~Nguyen and A.~Grover, ``{LICO}: Large language models for in-context molecular optimization,'' \emph{arXiv preprint arXiv:2406.18851}, 2024.

\bibitem{garnelo2018neural}
M.~Garnelo, J.~Schwarz, D.~Rosenbaum, F.~Viola, D.~J. Rezende, S.~Eslami, and Y.~W. Teh, ``Neural processes,'' \emph{arXiv preprint arXiv:1807.01622}, 2018.

\bibitem{lakshminarayanan2017simple}
B.~Lakshminarayanan, A.~Pritzel, and C.~Blundell, ``Simple and scalable predictive uncertainty estimation using deep ensembles,'' \emph{Advances in neural information processing systems}, vol.~30, 2017.

\bibitem{gal2016dropout}
Y.~Gal and Z.~Ghahramani, ``Dropout as a {B}ayesian approximation: Representing model uncertainty in deep learning,'' in \emph{International Conference on Machine Learning}.\hskip 1em plus 0.5em minus 0.4em\relax PMLR, 2016, pp. 1050--1059.

\bibitem{mixtabpfn}
D.~Xu, O.~Cirit, R.~Asadi, Y.~Sun, and W.~Wang, ``Mixture of in-context prompters for tabular {PFN}s,'' \emph{arXiv preprint arXiv:2405.16156}, 2024.

\end{thebibliography}

\appendix
\subsection{Detailed experiment results}
\label{appx:details}
Table \ref{tab:general} presents the performance of HARP and H-MoE on HLSyn, with and without pretraining on Iceberg. While the evaluation section focuses on comparing the G-TNP architecture and weak label augmentation, this experiment demonstrates that scaling a pretraining dataset improves model generalizability, regardless of the model architecture.

\begin{table}[h]
    \centering
    \caption{Iceberg’s dataset improve generalization}
    \label{tab:general}
    \begin{tabular}{l c c}
        \toprule
        Test MSE & From scratch & With pretraining \\
        \midrule
        HARP & 0.21 & 0.04 \\
        H-MoE & 0.12 & 0.10 \\
        \bottomrule
    \end{tabular}
\end{table}

\begin{figure}[H]
    \centering
    \subfloat[Pretraining on HLSyn]{
        \includegraphics[width=0.45\columnwidth]{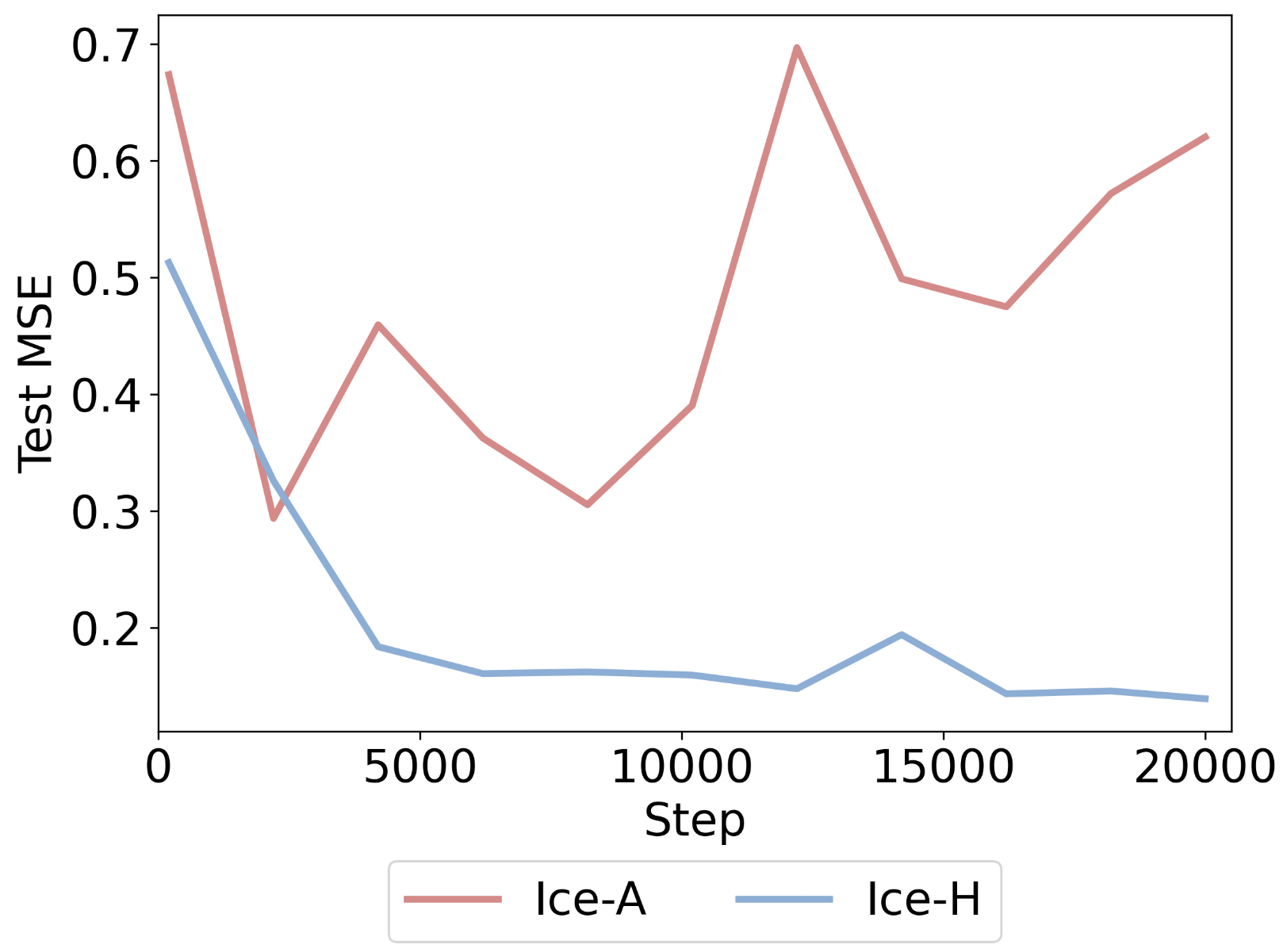}
        \label{fig:hlsyn_compare}
    }
    \hspace{1pt}
    \subfloat[Pretraining on Iceberg]{
        \includegraphics[width=0.45\columnwidth]{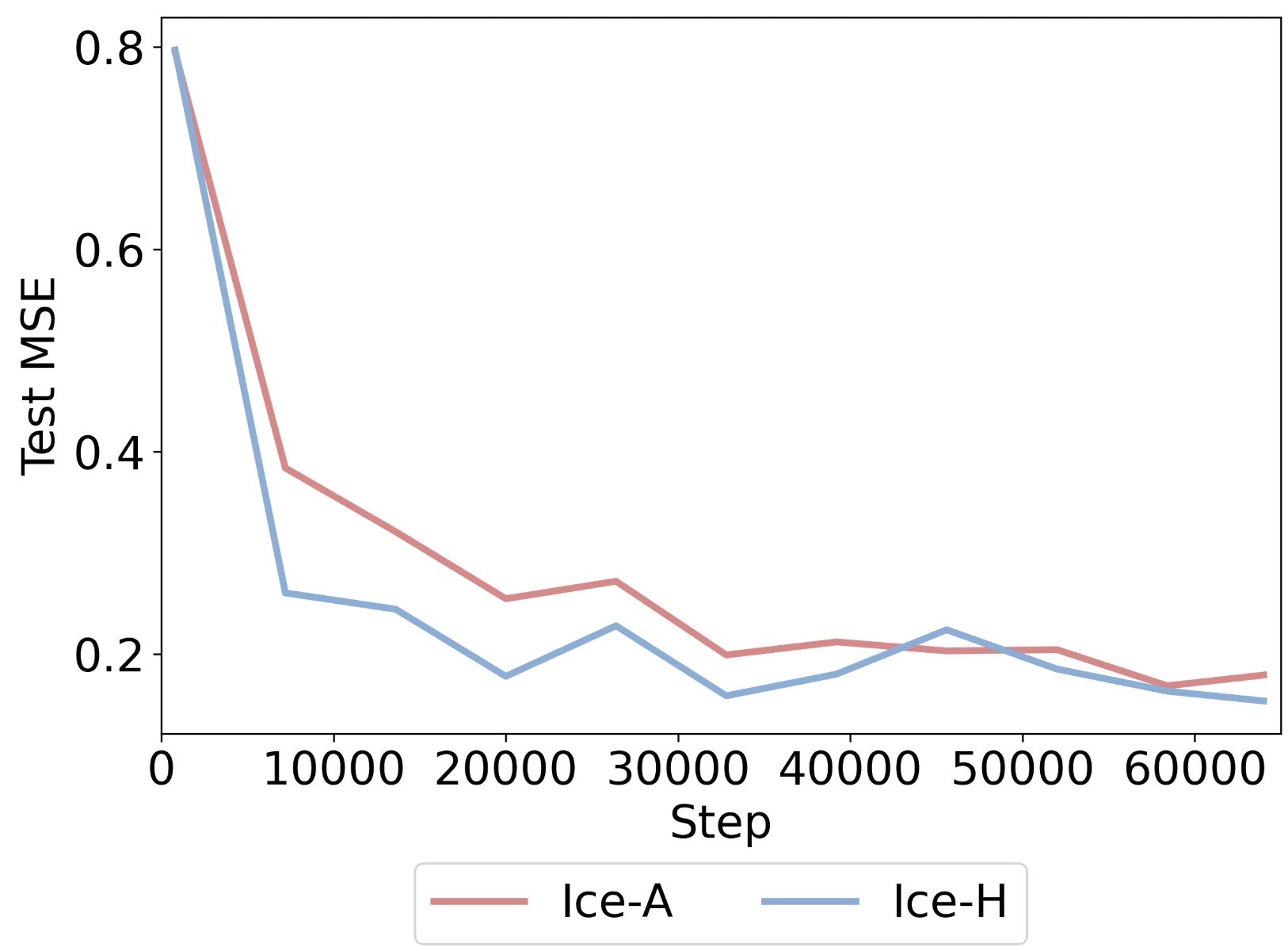}
        \label{fig:iceberg_compare}
    }
    \caption{\small Test loss curve when pretraining Ice-A and Ice-H with different datasets. The corresponding pretrained models are used in the evaluation of Sec. \ref{sec:eval_hlsyn} and Sec. \ref{sec:eval_iceberg}.}
    \label{fig:ah_loss}
\end{figure}

Figure \ref{fig:ah_loss} compares the testing loss curve when using either a subset of HLSyn or Iceberg as the pretraining dataset and a subset of HLSyn as the testing dataset, with and without weak label augmentation. We observe a significant accuracy boost when weak labels are added, particularly when the pretraining dataset is limited to a subset of HLSyn. However, as shown in Figure \ref{fig:iceberg_compare}, the testing loss difference between Ice-A and Ice-H is less significant, despite Ice-H outperforming Ice-A at the beginning and end of the training phase.

This aligns with the intuition that scaling actual labels is generally more effective than scaling potentially inaccurate weak labels. However, another key factor is the fidelity of the synthetic functions. When the training set is HLSyn, our synthetic function source --- the trained GNN models --- is relatively accurate. However, when switching the pretraining dataset to Iceberg, the GNN model does not achieve consistent modeling accuracy across all training programs, which may reduce the effectiveness of weak label augmentation through GNN-based synthetic functions.

As future work, we plan to investigate whether improving synthetic functions can further enhance performance. Another interesting research direction is to explore whether synthetic functions could directly contribute to optimization.

In Tables \ref{tab:break_iceberg_res_1}, \ref{tab:break_iceberg_res_2}, \ref{tab:break_best_1_hlsyn} and \ref{tab:break_best_1_real}, we present the result breakdown result of Figures \ref{fig:iceberg_res_1}, \ref{fig:iceberg_res_2}, \ref{fig:best_1_hlsyn} and \ref{fig:best_1_real}.

\begin{table*}[]
    \centering
    \caption{Result breakdown of Fig. \ref{fig:iceberg_res_1}}
    \label{tab:break_iceberg_res_1}
    \begin{tabular}{l c c c c c c}
        \toprule
        Benchmark & Ice-A & Ice-H & Ice-A-FT & Ice-H-FT & HARP & H-MoE \\
        \midrule
        2mm & 0.23 (0.02) & 0.29 (0.05) & 0.21 (0.04) & 0.19 (0.02) & 0.48 (0.12) & 0.35 (0.06) \\
        3mm & 0.14 (0.02) & 0.12 (0.04) & 0.10 (0.03) & 0.07 (0.03) & 0.20 (0.07) & 0.17 (0.02) \\
        covariance & 0.22 (0.03) & 0.26 (0.02) & 0.11 (0.02) & 0.11 (0.13) & 0.17 (0.03) & 0.15 (0.06) \\
        fdtd-2d & 0.25 (0.02) & 0.21 (0.04) & 0.21 (0.02) & 0.20 (0.06) & 0.24 (0.07) & 0.19 (0.05) \\
        gemver & 0.11 (0.02) & 0.10 (0.03) & 0.12 (0.02) & 0.12 (0.05) & 0.13 (0.02) & 0.10 (0.04) \\
        correlation & 0.20 (0.10) & 0.12 (0.11) & 0.03 (0.01) & 0.04 (0.04) & 0.60 (0.37) & 0.08 (0.04) \\
        fdtd-2d-large & 0.35 (0.10) & 0.23 (0.06) & 0.09 (0.01) & 0.11 (0.02) & 0.12 (0.05) & 0.12 (0.03) \\
        mvt-medium & 0.14 (0.01) & 0.13 (0.00) & 0.14 (0.01) & 0.10 (0.02) & 0.11 (0.03) & 0.13 (0.03) \\
        gemm-p-large & 0.08 (0.04) & 0.08 (0.05) & 0.04 (0.02) & 0.05 (0.03) & 0.10 (0.03) & 0.06 (0.02) \\
        atax-medium & 0.02 (0.01) & 0.01 (0.00) & 0.01 (0.01) & 0.00 (0.00) & 0.03 (0.01) & 0.02 (0.00) \\
        \midrule
        geomean & 0.14 (0.01) & 0.12 (0.02) & 0.08 (0.01) & 0.07 (0.02) & 0.16 (0.01) & 0.11 (0.01) \\
        \bottomrule
    \end{tabular}
\end{table*}

\subsection{Ablation study: scaling synthetic function diversity}
\label{appx:scale_f}

\begin{figure}[H]
    \centering
    \includegraphics[width=0.7\columnwidth]{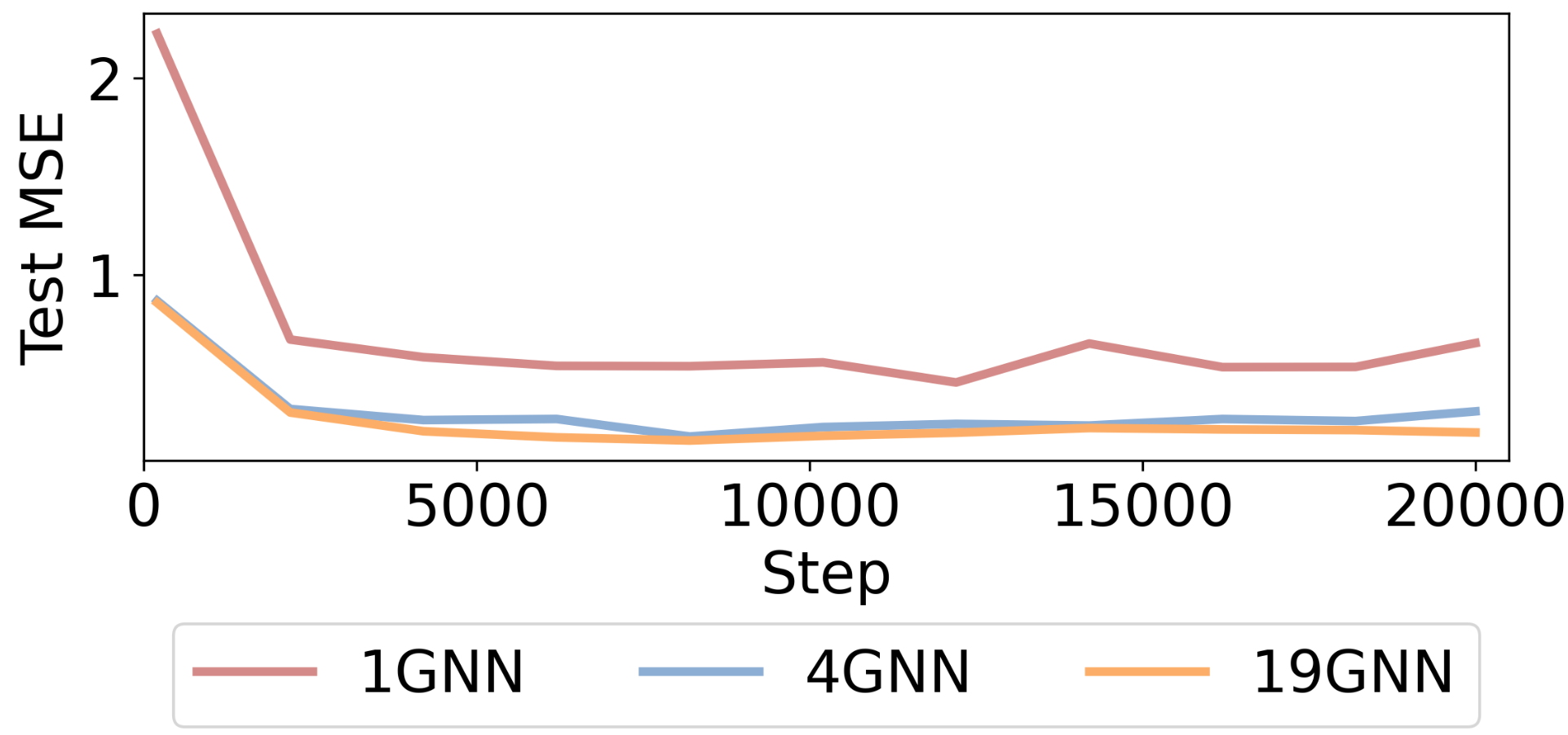}
    \caption{\small Model accuracy rises when generating weak labels with a more diverse set of synthetic functions. The GNNs are trained with different randomly initialized parameters, which has been found to effectively produce a calibrated set of models~\cite{lakshminarayanan2017simple}.}
    \label{fig:scale_syn}
    \vspace{-5pt}
\end{figure}

In Fig. \ref{fig:scale_syn}, we illustrate how the diversity of synthetic functions impacts the generalizability of the G-TNP model. The GNNs are trained on HLSyn as in Sec. \ref{sec:eval_hlsyn}. Our observations indicate that, even when the weak labels are accurate (Table \ref{tab:weak_label_acc}), better generalization is not simply achieved by training with more weak labels from unseen designs. Instead, weak label augmentation is effective only when the labels originate from multiple synthetic functions.

This aligns with the intuition that each synthetic function parameterized by $\theta_j$ provides a unique meta-learning task for the model. Learning from these diverse meta-learning tasks enhances the generalizability of the TNP model.

\subsection{Ablation study: comparing different sources of synthetic functions}
\label{appx:compare_syn}

\begin{figure}[h]
    \centering
    \includegraphics[width=0.85\columnwidth]{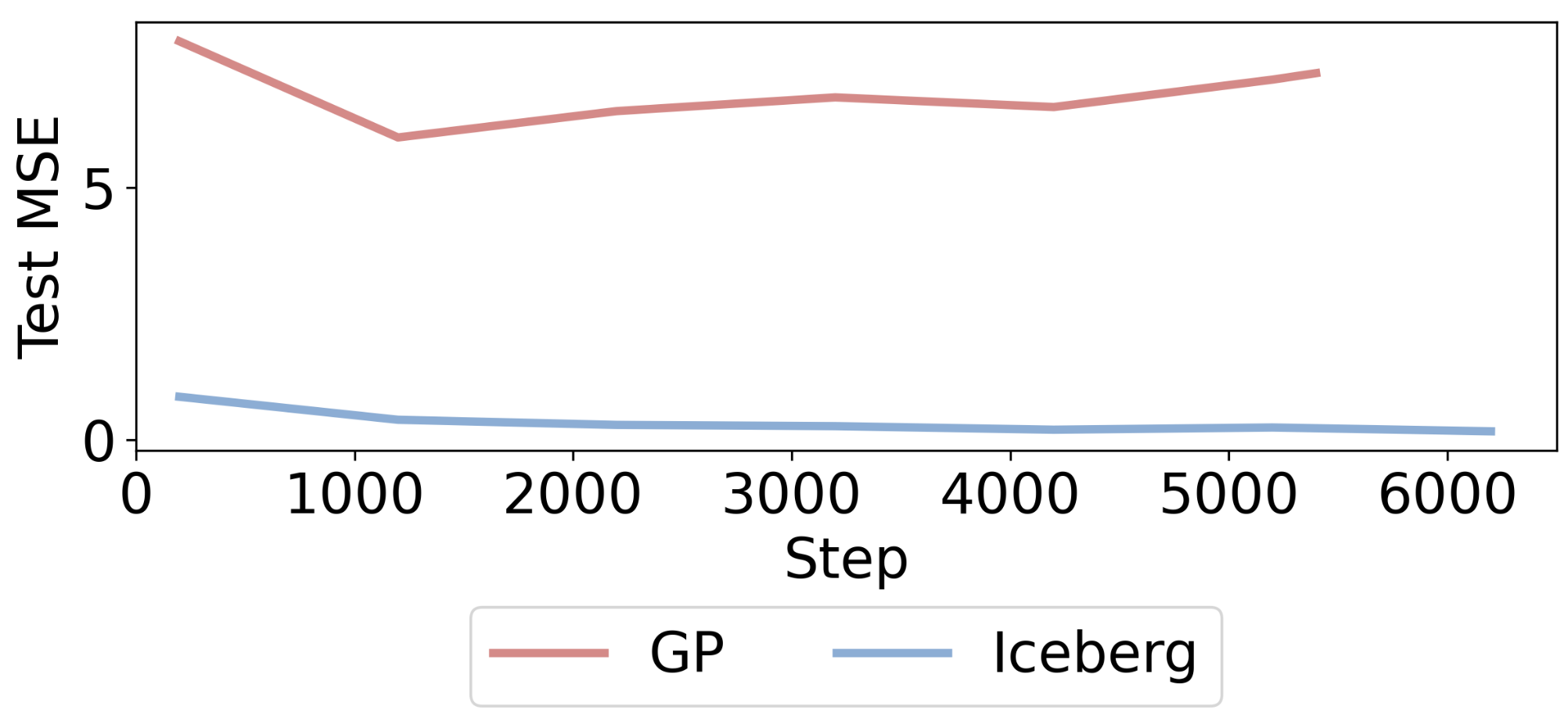}
    \caption{\small Comparison of Iceberg's weak label generation approach with previous methods. Iceberg uses an ensemble of trained GNNs, while GP represents a randomly initialized Gaussian Process. Iceberg's approach achieves significantly better results. The GP baseline is early stopped due to poor performance and signs of overfitting.}
    \label{fig:syn_compare}
\end{figure}

In Figure \ref{fig:syn_compare}, we present the testing loss curve using different sources of synthetic function for weak label generation. For this ablation study, we split the HLSyn dataset into training and testing sets as described in Sec. \ref{sec:eval_hlsyn}. To isolate the effect of weak labels, we train the model utilizing weak labels alone. The blue curve indicates the results when employing a Gaussian Process (GP) as the synthetic function. Following previous work, we implement GP with an RBF kernel:
$$\mathcal{GP}(0,\mathcal{K}), \quad\mathcal{K}(x,x')=\sigma^2\text{exp}\left(-\frac{(x-x')^2}{2l^2}\right)$$

We uniformly randomly sample the length $l$ from the range $[0.0, 7.0]$ and the scale $\sigma$ from the range $[1.0, 10.0]$.

The blue curve represents the results of Iceberg's weak label generation approach. We observe that our method significantly outperforms previous approaches, given the computational cost spent on training the GNN models.

While baseline methods could be further fine-tuned --- for example, by properly normalizing performance or sampling the mean of the GP --- we argue that our approach provides a straightforward and effective way to generate weak labels. Finding more computationally efficient sources for synthetic functions is left to future work.

Following prior work~\cite{expt,tabpfn}, it appears possible to train a larger and more robust model for various downstream tasks using a family of weaker functions. As future work, we plan to explore the potential and limitations of this approach.

\subsection{Background: pragma insertion for HLS}
\label{appx:pragma_insertion}
\begin{table}[]
\caption{\small Three types of HLS pragmas associated with each loop. CG: coarse-grained. FG: fine-grained.}
\label{tab:pragmas}
\centering
\begin{tabular}{c|ccc}
\hline
Name & Value Type & Available Options & Implication  \\
\hline
    PARALLEL & Discrete & factor=$<$int$>$ & CG/FG Parallelism \\
    PIPELINE & Categorical & mode=cg,fg & CG/FG Pipeline\\
    TILE & Discrete & factor=$<$int$>$ & Loop Tiling \\ 
\hline
\end{tabular}
\end{table}

\begin{figure}[bt]
\centering
\includegraphics[width=0.95\columnwidth]{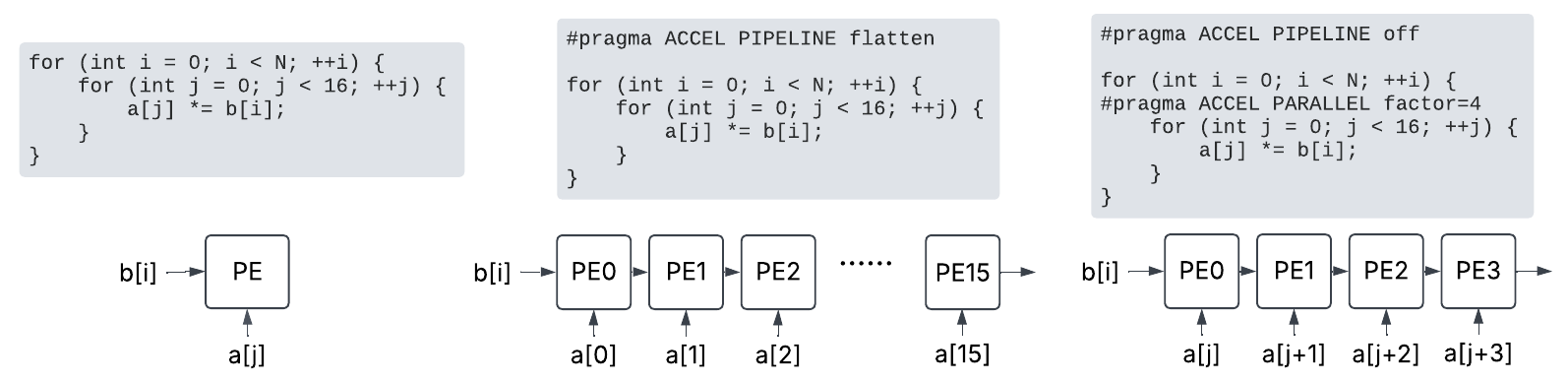}
\caption{Different combination of HLS pragmas means different circuit design. Here each PE multiply $a[j]$ with $b[i]$. Different pragmas will result in different latency and resource consumption.}
\label{fig:pragma_to_arch}
\end{figure}

Table~\ref{tab:pragmas} lists the three types of Merlin-HLS pragmas. In the design space, each loop includes these three types of pragmas. Figure~\ref{fig:pragma_to_arch} illustrates how different pragmas influence the resulting hardware architectures.

Notably, the pragma ``\#pragma ACCEL PIPELINE flatten" in the middle figure is equivalent to ``PIPELINE mode=fg" in Table~\ref{tab:pragmas}. It also corresponds to applying ``PIPELINE mode=cg" to loop $i$ and ``PARALLEL factor=16" to loop $j$, which results in the creation of 16 processing elements (PEs) that process $a[i]$ in parallel.

In contrast, in the right-most figure, applying ``PARALLEL factor=4" to the inner loop $j$ produces 4 parallel PEs instead.

\newpage

\begin{table}[h]
    \centering
    \caption{Result breakdown of Fig. \ref{fig:iceberg_res_2}}
    \label{tab:break_iceberg_res_2}
    \begin{tabular}{l c c c c}
        \toprule
        Benchmark & Ice-A & Ice-H & HARP & H-MoE \\
        \midrule
        knn & 0.20 (0.16) & 0.15 (0.05) & 2.05 (0.46) & 2.08 (0.49) \\
        att-3mm & 0.09 (0.06) & 0.08 (0.03) & 0.51 (0.14) & 0.46 (0.18) \\
        att-3mm-fuse & 0.01 (0.00) & 0.00 (0.00) & 0.45 (0.14) & 0.45 (0.15) \\
        conv2d & 1.00 (0.71) & 1.06 (0.89) & 4.26 (1.26) & 4.26 (1.26) \\
        optical-flow & 0.49 (0.12) & 0.39 (0.19) & 2.53 (0.13) & 0.32 (0.21) \\
        3d-rendering & 0.41 (0.06) & 0.30 (0.12) & 0.79 (0.06) & 1.07 (0.39) \\
        \midrule
        geomean & 0.17 (0.05) & 0.13 (0.02) & 1.26 (0.16) & 0.92 (0.10) \\
        \bottomrule
    \end{tabular}
\end{table}

\begin{table}[h]
    \centering
    \caption{Result breakdown of Fig. \ref{fig:best_1_hlsyn}}
    \label{tab:break_best_1_hlsyn}
    \begin{tabular}{l r r r r r}
        \toprule
        Benchmark & Ideal & Ice-A-FT & Ice-H-FT & HARP & H-MoE \\
        \midrule
        2mm & 4.4e3 & 7.2e3 & 7.2e3 & 5.6e3 & 1.6e6 \\
        3mm & 1.1e4 & 1.1e4 & 1.1e4 & 1.1e4 & 1.1e4 \\
        covariance & 1.6e4 & 1.6e4 & 1.6e4 & 6.7e5 & 1.9e4 \\
        fdtd-2d & 1.0e4 & 1.1e4 & 1.1e4 & 1.3e4 & 1.1e4 \\
        gemver & 1.0e4 & 1.7e5 & 1.2e4 & 1.7e5 & 1.7e5 \\
        correlation & 6.0e4 & 6.0e4 & 6.0e4 & 3.9e6 & 7.4e4 \\
        fdtd-2d-large & 1.3e6 & 1.3e6 & 1.3e6 & 1.3e6 & 1.3e6 \\
        mvt-medium & 4.1e4 & 4.1e4 & 4.1e4 & 4.1e4 & 7.1e4 \\
        gemm-p-large & 7.4e4 & 7.5e4 & 7.4e4 & 7.5e4 & 7.5e4 \\
        atax-medium & 8.3e4 & 8.3e4 & 8.3e4 & 8.3e4 & 8.3e4 \\
        \midrule
        geomean & 3.3e4 & 4.7e4 & 3.5e4 & 1.0e5 & 8.8e4 \\
        \bottomrule
    \end{tabular}
\end{table}

\begin{table}[h]
    \centering
    \caption{Result breakdown of Fig. \ref{fig:best_1_real}}
    \label{tab:break_best_1_real}
    \begin{tabular}{l rrrrr}
        \toprule
        Benchmark & Ideal & Ice-A-FT & Ice-H-FT & HARP & H-MoE \\
        \midrule
        knn & 7.2e7 & 7.2e7 & 7.2e7 & 1.1e8 & 7.2e7 \\
        att-3mm & 9.1e5 & 9.1e5 & 9.7e5 & 9.7e5 & 1.2e6 \\
        att-3mm-fuse & 2.7e7 & 2.7e7 & 2.7e7 & 2.7e7 & 2.7e7 \\
        conv2d & 1.2e8 & 1.2e8 & 1.2e8 & 3.4e9 & 1.7e8 \\
        optical-flow & \multicolumn{5}{c}{Resource Over Utilization} \\
        3d-rendering & 8.7e6 & 8.7e6 & 8.7e6 & 1.1e7 & 8.7e6 \\
        \midrule
        geomean & 1.8e7 & 1.8e7 & 1.8e7 & 4.0e7 & 2.0e7 \\
        \bottomrule
    \end{tabular}
\end{table}

\end{document}